\documentclass[a4paper]{llncs}

\usepackage{amssymb}
\usepackage{graphicx}

\usepackage{amssymb}
\usepackage{csquotes}
\usepackage{subfigure}
\usepackage{rotating}
\usepackage{algorithm}
\usepackage{footnote}
\usepackage{breakcites}
\usepackage[utf8]{inputenc}
\usepackage{url}

\urldef{\mailsa}\path|{linara.adilova, sven.giesselbach, stefan.rueping }@iais.fraunhofer.de| 

\newcommand{\keywords}[1]{\par\addvspace\baselineskip
\noindent\keywordname\enspace\ignorespaces#1}
\begin{document}

\mainmatter  

\title{Making Efficient Use of a Domain Expert's Time in Relation Extraction}

\titlerunning{Making Efficient Use of a Domain Expert's Time in Relation Extraction}

%
%
\author{Linara Adilova\and Sven Giesselbach\and Stefan R\"uping}
\authorrunning{Linara Adilova et al.}

\institute{Fraunhofer Institute for Intelligent Analysis and Information Systems IAIS\\
Schloss Birlinghoven, 53757 Sankt Augustin, Germany\\
\mailsa \\
}

%
%

\toctitle{Making Efficient Use of a Domain Expert's Time in Relation Extraction}
\tocauthor{Linara Adilova, Sven Giesselbach, and Stefan R\"uping}

\maketitle

%
%

\begin{abstract}
Scarcity of labeled data is one of the most frequent problems faced in machine learning. This is particularly true in relation extraction in text mining, where large corpora of texts exists in many application domains, while labeling of text data requires an expert to invest much time to read the documents. 
Overall, state-of-the art models, like the convolutional neural network used in this paper, achieve great results when trained on large enough amounts of labeled data. However, from a practical point of view the question arises whether this is the most efficient approach when one takes the manual effort of the expert into account. In this paper, we report on an alternative approach where we first construct a relation extraction model using distant supervision, and only later make use of a domain expert to refine the results. 
Distant supervision provides a mean of labeling data given known relations in a knowledge base, but it suffers from noisy labeling.
We introduce an active learning based extension, that allows our neural network to incorporate expert feedback and report on first results on a complex data set.
\keywords{relation extraction, convolutional neural networks, distant supervision, multi instance learning, interpretability, expert feedback}
\end{abstract}

\section{Introduction}
Nowadays, huge collections of textual data exist that do not only include interesting documents for humans to read, but can also be mined for interesting knowledge, which can be further stored in structured form. Examples include extracting general world knowledge from Wikipedia \cite{vrandevcic2014wikidata}, extracting knowledge about interactions of drugs, genes, and diseases from PubMed \cite{craven1999constructing,herrero2013ddi}, or definitions from arbitrary scientific publications \cite{augenstein2017semeval}. Currently, machine learning methods based on deep neural networks play an important role in the extraction of knowledge from texts, achieving top results on many benchmark data sets \cite{DBLP:journals/corr/SantosXZ15,lee2017semeval}. However, experience shows that deep learning works best in a supervised setting with a massive amount of labeled data. In practical applications, the effort to manually curate a large enough labeled data set is often prohibitively high, in particular in more specialized domains where highly trained domain experts are required. Even in cases where one is willing to invest a high manual effort, it may make more sense to extract the required knowledge completely manually because of the unfavorable ratio between effort and precision/recall for a supervised machine learning approach. 

In this paper, we address the problem of extracting relations from a large collection of documents with only negligible manual effort on the side of a domain expert. We target situations where currently knowledge extraction cannot be applied economically with respect to the manual effort required. Our approach is based on the idea of integrating the expert into the knowledge extraction process not before a deep learning method (as a mere labeling device), but instead by enabling the experts understanding of the extracted model and enabling him give high-level feedback on the results, that will be used to optimize the model.
%
%
A way of making use of knowledge - in the form of knowledge graphs - in relation extraction is distant supervision. In distant supervision knowledge that is stored in a knowledge graph is aligned with textual corpora. This yields a cheap way of automatically labeling new training data.


This paper is organized as follows: in the next section, we discuss background and related work on text mining methods with a focus on deep learning. Section \ref{sec:model} introduces our approach, both giving a detailed description of the interactive knowledge extraction process and the distantly supervised deep network that is applied in the intermediate steps. Section \ref{sec:experiments} gives first empirical results on the proposed approach. Section \ref{sec:conclusion} concludes and gives an outlook to future research.

\section{Background and Related Work}
In this section we shortly describe the theoretical background behind this paper and set it into the context of related work.

\subsection{Relation Extraction}
\label{subsec:re}
The task of Relation Extraction is about getting semantic meaning out of the sentences and texts that contain two entities mentions. This semantic meaning is later aligned with one of the pre-defined relations (so-called "fixed schema") or it can also be taken in its natural form as a new relation \cite{riedel2013relation}. Classical and the most used method for Relation Extraction is to get all possible linguistic characteristics of the raw textual data and then apply different kinds of classifiers on top of these constructed feature vectors \cite{zhang2006composite,culotta2004dependency}. With the development of Deep Learning this approach was also applied to Relation Extraction \cite{zeng2014relation,a27169dffe4642f78115676ce777c52c,DBLP:journals/corr/SantosXZ15}. Also, the question of finding ways to perform Relation Extraction without costly construction of training datasets was always getting a lot of attention. For example, Open Information Extraction \cite{banko2007open} performs the extraction without any human input. Furthermore, Distant Supervision was introduced \cite{Mintz:2009:DSR:1690219.1690287} as a way of utilizing existing structured data for obtaining training dataset without manual labeling of the examples.

\subsection{Ranking Convolutional Neural Networks for Relation Extraction}
\label{subsec:rcnn}
In \cite{DBLP:journals/corr/SantosXZ15}, a convolutional neural network for relation extraction is introduced.
The model consists of multiple layers, which we will quickly describe in this subsection:
\begin{enumerate}
\item \textbf{Word embedding layer:} Transforming words of the input sentence to the embeddings. Every word $w_i$ of the sentence transformed to $r^{w_i}$ that is a row of an 
embedding matrix $W^{wrd}$ for some fixed-size vocabulary.  
\item \textbf{Distance embedding layer:} Transforming distances between the words in the sentence 
  and marked named entities to the embedding vectors $wp_1$ and $wp_2$.
  This approach was introduced by \cite{zeng2014relation}.
\item \textbf{Embedding merge layer:} Concatenates the word embedding $r^w$ and corresponding 
  distance embeddings (to the first and to the second named entity) $wp_1$ and 
  $wp_2$ for every word $w$ in the input sentence into one vector.
\item \textbf{Convolutional layer:} Convolution is applied to windows of three embedding vectors
  with zero padding, so the size of the input is not changed after the layer application.
  The number of filters $d^c$ is 1000,
  where each value in one vector is a feature value for a specific   triplet of words.
  \item \textbf{Global maxpooling:} The maximal value is found for each filter. 
\item \textbf{Scoring dense layer} In order to classify relations the closeness of a sentence   representation to real valued vectors representing each of the relations that are learned during   the training process is  estimated.
  The scoring procedure is implemented as a dense layer without bias with weights matrix $W^{classes}$ consisting of the relations embeddings.
  \end{enumerate}

The objective function uses two of the resulting scores - one is the score, that was obtained for the 
correct relation according to the label of the example and the second score is one of the wrong 
relations scores.
Thus, the objective function is calculated as follows:
\begin{itemize}
  \item Get the score to the correct relation;
  \item Get the maximal score from the remaining wrong relations;
  \item Calculate the value of the loss according to the formula:
  $$L = log(1 + exp(\gamma(m^+ - s_{\theta}(x)_y^+))) + log(1 + exp(\gamma(m^- + s_{\theta}(x)_c^-)))$$
where $m^+$ ($m^-$) is a margin for the right (wrong) answer;
$\gamma$ is a scaling factor; $s_{\theta}(x)_y^+$ is a score for the right class; $s_{\theta}(x)_c^-$ is a score for the wrong class.
\end{itemize}

\subsection{Distant Supervision}
While deep learning architectures, such as the one discussed in the previous section, show excellent performance given enough labeled examples, in practice the problem arises that also abundant sentences can usually be found, labeling enough sentences is usually hard. Labeling is usually done either be crowd-sourcing \cite{angeli2014combining} by non-experts - which negatively influences the quality of the labels - or by experts, which in practice very much limits the amount of labeled examples one can generate.

In order to alleviate this problem, the approach of distant supervision \cite{Mintz:2009:DSR:1690219.1690287} has been proposed.
In order to apply this concept, along with the text corpus a structured knowledge base that contains examples of the desired relation ht base will be used to automatically generate examples of the relation by aligning the entities from the knowledge base with the text cee string matching or more complex entity recognition solutions can be used. 
Hence, distant supervision relies on the following two assumptions:

\begin{enumerate}
  \item For every triple $(e_1, e_2, r)$ in a knowledge base every sentence containing mentions for 
  $e_1$ and $e_2$  expresses the relation $r$
  \item Every triple that is not in the knowledge base is assumed to be a false example for a relation (even though the reason might be in the incompleteness of the knowledge base)
\end{enumerate}

Evidently, the better the knowledge base and text corpus fulfill these assumptions, the better one can expect the approach of distant supervision to work. In practice, it must be assumed that in addition to correct example sentences for the relation, additional noise is introduced. 


\subsection{Multi-instance Learning}
\label{subs:mil}
For coping with the noise introduced by distant supervision, we apply multi instance learning as described in \cite{zeng2015distant}. Multi-instance learning  was first introduced for drug classification
\cite{dietterich1997solving}.
	
In application for relation extraction, multi-instance learning will mean that we assume 
existence of at least one sentence containing the description of the relation from the 
knowledge base. So the set of all sentences that mention the same entity pair is considered as one bag and it has a 
corresponding label of the relation from the knowledge base. In order to apply neural networks 
for this bag-based training the maximal score example is chosen from the bag every 
time to fit the model, while all bags are shuffled from epoch to epoch.
This approach still looses a lot of possibly useful information obtained by distant 
supervision, but it serves as an initial step for possible improvement of the approach.

\subsection{Interpretability of a Deep Neural Network}
The interpretability of machine learning models, and in particular of deep networks, is currently receiving much attention. 
Several different directions for making a deep neural network model understandable to a domain expert have been proposed in the literature
\begin{description}
\item[Rule extraction:] Early approaches often focused on the extraction of rules or other understandable representations from neural networks, e.g. \cite{thrun1995extracting}. 
However, for complex data such as texts, and complex models, there is rarely a concise understandable model that summarizes the whole network, hence these approaches have fallen out of interest.
\item[Relevance propagation and feature weights:] for each individual prediction, it is possible to trace which part of the model in each layer was how relevant for taking the final decision \cite{bach2015pixel,binder2016layer}. 
Taking the relevance propagation back to the level of the input features, this gives an importance for each input feature.
\item[Local Approximations:] each prediction of the classifier can be approximated locally by simpler model \cite{ribeiro2016should}. 
The understandability of the approximating model can be guaranteed by using a simple model class, e.g.~a linear model.
\item[Joint Models:] in certain cases, it is also possible to construct models that both give a highly accurate prediction plus a reason for the prediction. E.g.~in \cite{lei2016rationalizing}, 
together with the model an explanatory phrase is extracted.
\item[Instance-based models:] these methods explain a classifier by means of representative instances, such as prototypes (typical well-classified instances), or critics (typical mis-classified instances) \cite{kim2016examples}. 
\end{description}

In general, generic method for classifier interpretation are hard to adapt to models working on text data, since it is a complex data type of low structure.
In the case of the approach of \cite{DBLP:journals/corr/SantosXZ15}, which is used in this paper, the authors suggest to extract representative trigrams from the text. The idea is based on relevance propagation, however make use of the property that in this model the convolutional layer is applied to three embeddings of consecutive words each. Therefore, for each trigram in the original sentence, its relevance for the predicted relation can easily be traced back.
More precise, representative trigrams can be obtained from 
 the sentences of the dataset by measuring the value that each of the trigrams in a sentence contribute to the  correct class score. The value is simply a sum of all score positions that are traced back to that 
 specific trigram. This method is very similar to the one mentioned in 
 \cite{craven1999constructing} when the most valuable words were extracted in order to 
 have an insight into the concept learned by the model.

\section{Model Description}\label{sec:model}
For our experiments we first implemented the ranking convolutional neural network as described in \cite{DBLP:journals/corr/SantosXZ15}. We added multi instance learning to cope with noise from distant supervision as well as a feedback loop for domain experts, that lets them improve the performance by evaluating the most representative trigrams for each relation type. 

\subsection{Expert Feedback for Training Data Curation}
An obvious problem with data set creation via distant supervision is that it adds training sentences that do not represent the relation they were sampled for. Instead of letting experts review all of those samples we propose an approach in which the expert reviews not each sentences but rather the concepts that the neural network learned for reach relation.

The representative trigrams that our model learns for each relation class can be regarded as its concepts for each relation. If the concepts make sense, our model most likely has achieved a good understanding of the relation from the distant supervised data. If not, assuming that our model is appropriate, the training data is probably not representative for the relations.

\begin{figure}[H]
\centering
\includegraphics[width=0.7\textwidth]{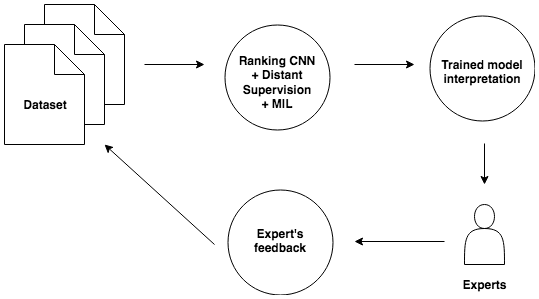}
\caption{Active learning approach diagram.}
\label{fig:act-learn-diagr}
\end{figure}

We propose the following workflow for dataset creation and model improvement, displayed in Figure \ref{fig:act-learn-diagr}:
\begin{enumerate}
\item Acquire/Construct a knowledge base with representative facts for relations of interest and text corpora that contain information about the entities and relations of interest.
\item Align the knowledge of the knowledge base with the text corpora and train a Ranking CNN with multi instance learning on it.
\item Extract representative trigrams for each relation class.
\item Show the representative trigrams to experts. Use the trigrams for performance evaluation of the model. Let the expert analyze what mistakes happened and let them filter out non-representative trigrams.
\item Filter out the training sentences for the relations which contain non-representative trigrams and start the process again with the redefined training set.
\end{enumerate}

Our assumption here is that the sentences that contain non-representative trigrams are the ones that confuse the network the most. Removing them should at least lead to better precision of the models. The analysis of the trigrams that the network deems as most important can yield many insights on the causes of bad accuracies.

\begin{figure}[H]
\centering
\includegraphics[width=0.7\textwidth]{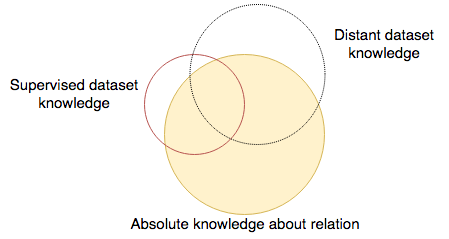}
\caption[Relation of supervised and distantly supervised knowledge]{Concept of knowledge about a specific relation contained in supervised and distantly supervised datasets.}
\label{fig:dist-superv-knowledge}
\end{figure}
If we imagine the concept of a relation as a set of knowledge, than ideally a supervised datasets capture the whole knowledge. In practice this is hardly feasible. An expert would have to label as many relevant sentences as possible, to capture the variance of the whole knowledge about the relation. Real supervised datasets rather represent a subset of the knowledge about a relation and also some noise e.g. because of wrong labeling. If we add a distantly supervised dataset we will most likely capture 3 different subsets of the overall knowledge: (1) knowledge or noise that is already included in the supervised dataset, (2) new knowledge and (3) new noise. With our proposed workflow we hope to reduce the size of the third subset, namely the newly introduced noise. This idea is reflected in Figure \ref{fig:dist-superv-knowledge}.

It is important to note that knowledge that is not reflected in the supervised training set will most likely also not be reflected in the supervised testing set. This leads to the assumption that we will underestimate the performance of our distantly supervised models when evaluating on test sets of supervised data sets.

\section{Evaluation}
\label{sec:experiments} 
We now evaluate the model architecture. We compare supervised training against distantly supervised training and highlight benefits and downsides of both approaches. We investigate the influence of multi-instance-learning and joint supervised and distant supervised learning. Lastly we evaluate the effect of the expert feedback on the quality of the resulting model.

\subsection{Data Sets}
\subsubsection{SemEval Task8}
We first evaluate our dataset on the SemEval task8 dataset. This dataset was originally used by \cite{DBLP:journals/corr/SantosXZ15} and we use it for model validation and comparison. The dataset contains nine bidirectional relation types and the "Other" class, that includes different relations, not included in the main ones. Hence there are 19 different relation classes. The sample sentences were manually collected from the web and annotated in three rounds, ensuring that all annotators agree on the label given to the sentence. 


\subsubsection{The KBP37 Dataset}
The KBP37 dataset\footnote{\url{https://github.com/zhangdongxu/kbp37}}, as it was called in the paper \cite{DBLP:journals/corr/ZhangW15a}, is a revision of MIML-RE annotation dataset from \cite{angeli2014combining}, that was build from a subset of Wikipedia articles by manual annotation. The benefit of KBP37 is that it is alignable with the Wikidata\footnote{https://query.wikidata.org/} and KBP-slot-filling datasets\footnote{\url{https://nlp.stanford.edu/software/mimlre.shtml}}. The following changes were made to the KBP37 datasets by the authors of \cite{DBLP:journals/corr/ZhangW15a} to adapt it to the description of the SemEval task8:
\begin{itemize}
  \item Added direction to the relations, i.e. 'per:employee-of(e1,e2)' and 'per:employee-of(e2,e1)' 
  instead of simply 'per:employee-of'. This is done for all the relations except for 'no-relation'
  \item Balance the dataset, to exclude the relations that have less than 100 examples for each of 
  the directions. Also $80\%$ of 'no-relation' examples are discarded
  \item After that examples are shuffled and split into three parts, $70\%$ for train, $10\%$ for 
  development and the rest for testing.  
\end{itemize}
After all modifications the dataset consists of 18 directional relations and one "no-relation" class, that will result in 37 classes for recognition. The dataset is more complex than the SemEval task8 dataset. It contains longer sentences (almost twice as long as the longest in SemEval) and it also has multi-relational pairs, making it closer to the real world problem of relation extraction but also more difficult to solve. Also it can be observed that the relations and entities in this dataset are more 
specific. Most of the entities in the dataset are either names of persons or companies. The relations are very specific, e.g. there are three different classes for placement of 
headquarters of a company. One for city, state and country.

One more important aspect of the dataset is that human labeling is error prone. Thus there are also very imprecise examples. Here are two examples for the alternate names class:
\\
\\
\textit{It was because of $<e1>$Abu Talib$</e1>$ 's ( a.s. ) good fortune that apart from $<e2>$his$</e2>$ ancestral services and prestige he also inherited from sons of Ismail ( a.s. ) high status and courage. \textbf{per:alternate-names(e2,e1)}}
\\
\\
\textit{The discography of $<e1>$ Billie Piper $</e1>$ ( as known as $<e2>$ Billie $</e2>$) an English pop music singer consists of two studio albums two compilation albums and nine singles.
\textbf{per:alternate-names(e2,e1)}}
\\
\\
In the second sentence we have a well labeled example for the class. The first sentence though is hardly an alternate name. It is rather an example for an anaphora resolution task. Such ambiguous labeling will make the classification task even more difficult as it is not obvious even for human annotators why both examples should belong to the same class.


\subsubsection{Knowledge Bases for Distant Supervision}
As knowledge bases for distant supervision we used both relational pairs from MIML-RE\footnote{\url{https://nlp.stanford.edu/software/mimlre.shtml}}, i.e. from TAC KBP, and Wikidata\footnote{https://query.wikidata.org/}.
Wikidata \cite{vrandevcic2014wikidata} is a crowd-sourced knowledge base. Its users collaborate on filling it with facts, but they also collaborate on validating the data and updating the scheme of the knowledge base. 
The TAC KBP data is from a knowledge base population task by the Text Analysis Conference, with the goal of discovering information about entities and incorporate it into a Knowledge Base. For relational facts alignment the knowledge base of the Stanford Natural Language Processing group was used\footnote{\url{https://nlp.stanford.edu/software/mimlre.shtml}}.  

Knowledge base relations were aligned with the New York 
Times corpus\footnote{\url{https://catalog.ldc.upenn.edu/ldc2008t19}}. 
The amount of entity pairs for each of the relations varies a lot - from less than 1000 to more than 
50000. In order to create an artificial "Other" class we chose the relations "per:religion", "per:children" and "org:political/religious-affiliation". When investigating the entity pairs from MIML-RE we found them to be not very accurate. An example being an entity of the type "person-name" that contains only a single letter. In order to minimize noise effect of these pairs, 
entity pairs from Wikidata were added to the knowledge base. Wikidata contains less matching data for the 
corresponding relations, but the relations are more precise. 
We additionally cleaned entity pairs by removing the ones containing one-letter entities or names consisting only of capital letters with dots.

To align the knowledge bases with our textual corpus, we simply matched the strings of the entity names with the texts. If a sentence includes both entities which are part of a relation, we used it as sample for the relation.

\subsection{Supervised Training Evaluation}

To validate the correctness of the implementation of the ranking convolutional neural network described in Section \ref{subsec:rcnn}, it was tested on the test set of SemEval2010 Task8 dataset and KBP37 test dataset. The scores we achieved is compared to other scores in Table \ref{tab:test-superv-general}. We can conclude, that the model achieves comparable quality to the reference model and our implementation seems to be correct. We also notice, that the results achieved with our CNN are higher than with the Recurrent Neural Network from \cite{DBLP:journals/corr/ZhangW15a}.

\begin{table}
  \begin{center}
 \begin{tabular}{ | c | c | c | }
    \hline
    Classifier & SemEval2010 & KBP37 \\ \hline
    CR-CNN \cite{DBLP:journals/corr/SantosXZ15} & 84.1 & - \\ \hline
    RNN \cite{DBLP:journals/corr/ZhangW15a} & 79.6 & 58.8 \\ \hline
    Supervised Ranking CNN & \textbf{84.39} & \textbf{61.26} \\ \hline
    \end{tabular}
\caption{F1-scores for testing datasets.}
\label{tab:test-superv-general}
\end{center}
\end{table}


\subsection{Distant Supervision Evaluation}

The results of training the network in various ways with distant supervision can be seen
in Table \ref{tab:dist-gen-res}. For comparison we also add the results of supervised training.

\begin{table}
  \begin{center}
 \begin{tabular}{ | c | c | c | c | c |}
    \hline
    Experiment & P & R & F1 & Manual Effort \\ \hline
    Supervised training & \textbf{67.74} & \textbf{57.88} & \textbf{61.26} & 17638\\ \hline
    Distantly supervised training & 50.71 & 45.24 & 43.81 & \textbf{0}\\ \hline
    Distantly supervised + MIL & 51.82 & 46.61 & 45.40 & \textbf{0}\\ \hline
    \end{tabular}
\caption{Precision, Recall, F1-scores and manual effort (number of sentences the expert has to label) for distantly supervised training evaluation.}
\label{tab:dist-gen-res}
\end{center}
\end{table}

While distant supervision performs worse than supervised training - which was to be expected - the results are usable in practice. In particular, they are significantly higher results than random assignment (with 37 classes, the
F1-score for random assignment would be around 0.2\%). A publicly available knowledge base and appropriate text corpora can hence serve for the automatic creation of a training set for a neural network tackling the task of relation extraction. 

Moreover, in the context of the task to continuously extract new knowledge from newly published texts under a constraint budget of manual intervention, this approach is more appealing than both manual extraction and supervised training.
We can quantify the savings on the side of the expert by evaluating how many sentences an expert would have to read in each of the settings: 
for the KBP37 dataset, the number of sentences in the testing set is 3403 and in order to get relations from them the experts should fully comprehend all the information. Moreover, with the manual approach this should be done for all new texts again. The  manually supervised approach would require full comprehension for creating the training dataset that is 17638 sentences and later the experts would check the obtained results (1969 sentences). For the distant supervision on the other hand, all that is required is a result check that is around 1586 sentences and it can be repeated continuously to get all the relations.

The second observation we can make is that multi-instance learning has a slight positive effect on the performance. Multi-instance learning improved the results in every experiment by almost 2\%. Multi-instance learning did improve precision and 
recall simultaneously.



It is also important to notice, that the supervised training and testing datasets are tightly coupled and they will have common context and common biases. Thus evaluating the distantly supervised model on the existing testing
dataset might not be an objective choice. There exist other ways to evaluate the results of distant supervision, for example, as done in \cite{Mintz:2009:DSR:1690219.1690287}, but they would not show a realistic comparison to the  supervised results. 



\begin{figure}[!h]
\centering
\includegraphics[width=.80\textwidth]{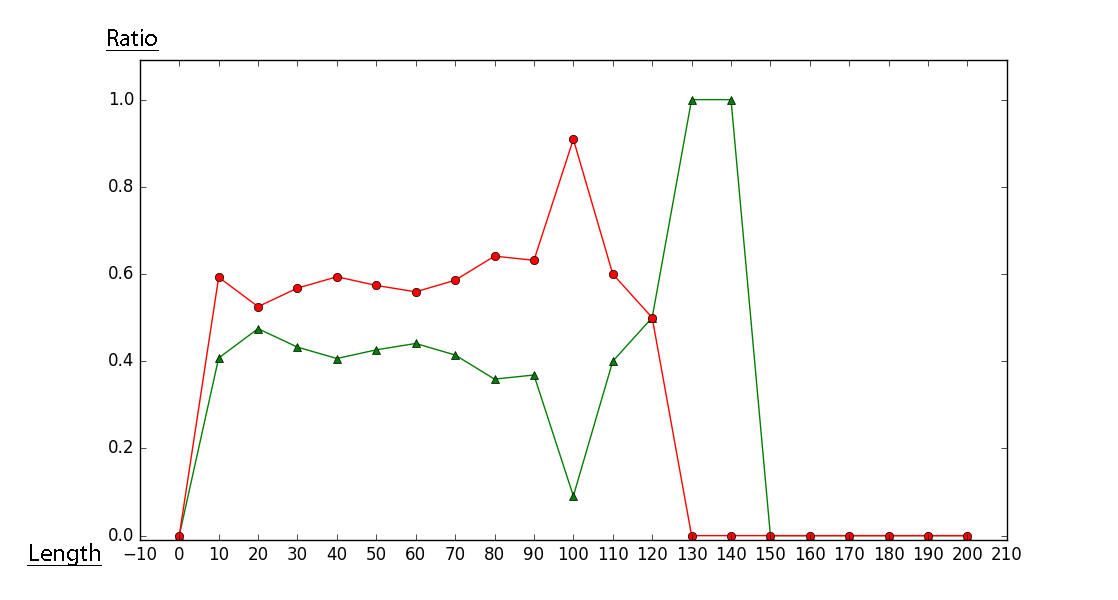}
\caption{Correlation of amount of correct and wrong answers with sentence length. Number of correct answers (green) and wrong answers (red) is normalized by the overall amount of examples of specific length.}
\label{fig:dist-depend}
\end{figure}

Furthermore, we investigate the dependency between the performance of the approach and the complexity of the sentences.
The dependency can be seen in the Figure \ref{fig:dist-depend}. Spikes around the large values of length are not representative, as the number of the examples there much smaller (3-5 sentences). For all other values, with higher length the number of errors grows and the number of right answers drops.
Any distantly supervised dataset will always be characterized by longer sentences on average, so this aspect should be taken into 
account when the dataset is constructed. For example, sentences longer than some limit can simply not be included in the final set of training examples.

\begin{table}
\centering
\begin{tabular}{ | c | p{7cm} | }
\hline
\textbf{org:founded-by} & \textit{founder of the; open society institute; fox broadcasting company; ethical treatment of}\\\hline
\textbf{per:alternate-names} & \textit{known as dwight; known as dj; known as milli; known as matthew;  real name is; real name was; , mimi smith; name was selena}\\\hline
\textbf{org:members} & \textit{soccer league milwaukee; american league boston; national league colorado; football league saskatchewan; midwest league burlington;  hockey league .; football league and; basketball league ,; football league 's; soccer league , }\\\hline
\textbf{org:top-members/employees} & \textit{said gene russianoff; chief operating officer; , managing director; , chief executive; chief executive of; sony pictures entertainment; executive vice president; , vice president}\\\hline
\textbf{per:countries-of-residence} & \textit{england .; france .; states .; australia .; united states ,; philharmonic .; ) italy :; the like ,}\\\hline
\textbf{org:founded} & \textit{, 2000 .; the 1980 's; , 2001 ,; in 1997 ,; in 1996 ,;  , 2001 ,; , 2000 ,; in 1997 ,; in 1999 ,; in 1998 ,}\\\hline
\textbf{org:subsidiaries} & \textit{, a subsidiary; high school in; the walt disney; a division iii; the university of; high school ,; department stores company; general motors corporation}\\\hline
\textbf{per:employee-of} & \textit{( columbia ); secretary of state; senator daniel inouye; senator sam brownback; ( columbia ); ( interscope ); ( atlantic ); blue note label}\\\hline
\textbf{per:country-of-birth} & \textit{england .; states .; france .; africa .; united states ,; united states in; , england ,; united states attorney}\\\hline
\textbf{per:cities-of-residence} & \textit{los angeles ,; los angeles band; revved-up vancouver outfit; in london ,; city .; paris .; los angeles ,; angeles .}\\\hline
\textbf{org:alternate-names} & \textit{states department of; california , los; and municipal employees; the university of; known as dwight; known as dj; known as milli; known as matthew}\\\hline
\textbf{org:country-of-headquarters} & \textit{the university of; states .; japan .; germany .; in london ,; york city ,; arbor , mich; cambridge , mass}\\\hline
\textbf{org:stateorprovince-of-headquarters} & \textit{university school of; the university of; , ohio ,; university .; the university of; university in tokyo; life insurance company; institute of technology}\\\hline
\textbf{per:spouse} & \textit{benazir bhutto ,; brad pitt and; david lynch ; starring david arquette; and her husband;by richard gere; director herbert ross; starring michael douglas}\\\hline
\textbf{org:city-of-headquarters} & \textit{in london ,; york city ,; arbor , mich; cambridge , mass; the university of; hill , n; arlington , va; city .; }\\\hline
\textbf{per:stateorprovinces-of-residence} & \textit{california .; york .; of california at; florida .; new york ,; new york city; in california ,; new york times}\\\hline
\textbf{per:title} & \textit{) film review; director of; ) television review; the actor who; the director of, this film is; director of; prime minister ,; the director ,}\\\hline
\textbf{per:origin} & \textit{of american art; the american artist; the american painter; 20th-century american art; american art ,; american art .; american academy of; american art at; french mathematician ,}\\\hline
\end{tabular}
\caption{Representative trigrams.}
\label{rep_trigrams_experts}
\end{table}

To inspect the model in more detail, we extracted the representative trigrams for each class, see Table \ref{rep_trigrams_experts}.
A first immediate finding of looking at the trigrams was that many of them make sense but tend to include the names of entities and might hence even overfit to the names in the training set. For example, for the relation \textit{org:founded}, it is obvious that the concrete years should be replaced by a placeholder.

\subsection{Using Expert Feedback}
We have seen that we can construct a useful data set for relation extraction using  distant supervision and multi instance learning. Now we want to evaluate whether feedback from experts about the concepts learned by our model can be used to improve the quality of our dataset and the model.
For this experiment we used the model trained on the distantly supervised data set with the relations from KBP37 and sentences from the New York Times corpus.

To evaluate the approach of integrating expert feedback to improve the model, we conduct the following experiment: from the representative trigrams of Table \ref{rep_trigrams_experts}, we select nonsensical trigrams plus trigrams that are too specific, e.g.~overfit on specific names. Sentences matching those trigrams are removed from the training set, as they are suspected to introduce too much noise, and the model is trained again on the filtered data set. Table \ref{tab:dist-gen-res} shows the results, with classes where the F1-score changes by less than $1.00$ between the initial and filtered run are excluded because of space constraints.

\begin{table}
  \begin{center}
 \begin{tabular}{ | c | c | c | c | }
    \hline
    Class & F1-score initial & F1-score filtered & New sensible\\
    &&&trigrams \\ \hline
    org:founded-by & {\bf 19.30} & 13.33 & 2 \\ \hline
    org:members & {\bf 14.86} & 13.64 & 0 \\ \hline
    org:top-members/employees & {\bf 44.77} & 40.49 & 1 \\ \hline
    per:alternate-names & 24.72 & {\bf 31.17} & 1 \\ \hline
    per:cities-of-residence & 52.82 & {\bf 54.73} & 0 \\ \hline
    per:countries-of-residence & 9.33 & {\bf 13.20} & 0 \\ \hline
    per:country-of-birth & 25.83 & {\bf 26.86} & 0 \\ \hline
    per:employee-of & {\bf 43.39} & 39.91 & 3 \\ \hline
    per:spouse & {\bf 43.56} & 36.00 & 1 \\ \hline
    per:stateorprovinces-of-residence & 43.95 & {\bf 45.49} & 1 \\ \hline
    per:title & 87.45 & {\bf 87.55} & 2 \\ \hline
    \end{tabular}
\caption{Changes in F1-score after filtering examples with non representative trigrams in them. Training is performed without Multiple Instance Learning. Classes with difference in F1 smaller that $1.00$ excluded due to space constraints.}
\label{tab:initial-vs-filtered}
\end{center}
\end{table}

In detail, the following effects can be seen in the trigrams:
\begin{description}
\item["per:cities-of-residence":] all the trigrams contained names of the cities. Here, even after filtering all the new trigrams contain only city names. 

\item["per:countries-of-residence":] a lot of non representative trigrams were filtered. As a result, network started to concentrate more on the persons names in the form of "johan anderson of", "vanessa gusmeroli of". 



\item["org:founded-by":] most of the trigrams included companies names, so they were filtered out. This allowed to obtain other trigrams such as "clifford noble opened" or "dick clark productions" but it worsened the overall score as previously learned company names were not taken into account anymore. 

\item["org:members":] the data to the information about sport leagues and the trigrams contained only leagues names both before and after filtering.  

\item["per:stateorprovince-of-residence":] performance improved as it started to learn also constructions like "pete domenici of".


\item["org:top-members/employees":] had a lot of persons names in its trigrams. So, filtering them out again affected overfitting of the network, but it allowed to get such trigrams as "editor of the" for example.

\item["per:alternate-names":] filtering out trigrams with names allowed to get "real name was" for example without loosing "real name is" and "known as dwight". In this case the network started to see really good constructions.

\item["per:country-of-birth":] after filtering started learning persons names more, that helped it to give better results.

\item["per:employee-of":] overfits to companies names. Filtering trigrams allowed to get "of state" and "former defense secretary" but worsened the result because it does not make conclusion by the company name anymore.


\item["per:spouse":] the relation has a lot of training examples with celebrity names. All of such trigrams were filtered out. It allowed to learn at least "her husband," leaving all the others names again.

\end{description}

At first glance, the results may look unconvincing: results improve for 5 relations, but get worse for 5 relations. However, looking at the trigrams before and after the filtering
the following two observations can be made:
\begin{enumerate}
\item {\bf Performance is mostly influenced by overfitting on entities:} it is clear from looking at the trigrams, that very often concrete names of cities, persons, or organizations are learned, which is not a desired behaviour. Because of the random training and test split, very often these entities occur in both training and test data, such that good results are obtained still. Removing these trigrams has a negative effect in most cases, as no more general relations can be learned. Interestingly, in some cases removing non-sensical trigrams allows the network to identify even more concrete entities which improves the results, e.g.~in the case of "per:country-of-birth".
\item {\bf More sensible trigrams improve the results:} some examples, e.g.~the relation "per:alternate-names" or "per:stateorprovince-of-residence" show improved results with more sensical trigrams.
\end{enumerate}
In summary, it might be meaningful for the expert to make the decision on which trigrams to include based on a comparision of the trigrams both before and after the filtering: in the case where no more meaningful trigrams are found, it might make sense to conclude that no general model can be found and not filter the overfitted trigrams afterall.

\section{Conclusion and Outlook}\label{sec:conclusion}
Despite the many successes of deep learning in relation extraction, for many practical problems, the availability of labeled data is the main limiting factor. Due to the complexity of the knowledge that is to be extracted from the texts, supervised approaches need many more examples than what usually is available in practical applications. 
In this paper we explored possibilities to make use of a domain experts knowledge in a more efficient way than using him as a labeling device.

It has been shown that distant supervision, in combination with multi-instance learning, is a meaningful method for relation extraction and well surpasses both manual information extraction and state-of-the-art supervised approaches when performance in relation to manual effort is concerned. The necessary effort by the domain experts can in this case be constrained to the identification of a meaningful structured database for generating distantly supervised examples.

An analysis of distant supervision and multi-instance learning in the specific case of the KBP37 dataset showed that the quality of the attainable results can be limited effects of overfitting on specific entities. We have shown that in this case the domain expert can contribute by inspecting the predictions made by the deep model on the level of representative trigrams.
With the insight gained, the expert can contribute to improve the quality of the model by removing examples that were wrongly labeled by distant supervision, or giving input on pre-processing steps that may help the generalization ability of the model.

Future work will aim at a more in-depth evaluation of the approach. 
Our hypothesis is that the presented approach will be more effective in the case of more specialized relations and in-depth knowledge, for example in the case of medical texts. 
Finally, obviously representative trigrams are only a very coarse tool for making the model more understandable.


\paragraph{Acknowledgements:}
This work upon which this paper is based was supported by means of the Bundesministerium f\"ur Bildung und Forschung (F\"orderkennzeichen 031L0025C).




\bibliography{literature}
\bibliographystyle{apalike}

%
%


\end{document}